\documentclass[runningheads]{llncs}
\usepackage[T1]{fontenc}
\usepackage{graphicx}
\usepackage{amsmath}
\usepackage{amssymb}
\usepackage{url}
\usepackage{breqn}
\usepackage{times}
\usepackage{epsfig}
\usepackage{cite}
\usepackage{amsmath,amssymb,amsfonts}
\usepackage{algorithmic}
\usepackage{graphicx}
\usepackage{textcomp}
\usepackage{xcolor}
\usepackage{floatrow}
\usepackage{blindtext}
\usepackage{colortbl}

\begin{document}
\title{PASTA: PArallel Spatio-Temporal Attention \\ with spatial auto-correlation gating \\ for fine-grained crowd flow prediction}
\titlerunning{Parallel Spatio-Temporal Attention with Spatial Auto-Correlation Gating}
%


\author{Chung Park \inst{1,3} \and
Junui Hong \inst{1,3} \and
Cheonbok Park \inst{2} \and
Taesan Kim \inst{1} \and
Minsung Choi \inst{1} \and
Jaegul Choo \inst{3} \thanks{Corresponding author}}

\authorrunning{Park et al.}
\institute{SK Telecom, Seoul, Republic of Korea \\
\email{\{skt.cpark, skt.juhong, ktmountain, ms.choi\}@sk.com}\\
\and
Naver Corp, Seongnam, Republic of Korea\\
\email{cbok.park@navercorp.com}
\and
Kim Jaechul Graduate School of AI, KAIST, Daejeon, Republic of Korea \\ \email{\{cpark88kr, secondrun3, jchoo\}@kaist.ac.kr}}

\maketitle 
\begin{abstract}
Understanding the movement patterns of objects (e.g., humans and vehicles) in a city is essential for many applications, including city planning and management. 
This paper proposes a method for predicting future city-wide crowd flows by modeling the spatio-temporal patterns of historical crowd flows in fine-grained city-wide maps. 
We introduce a novel neural network named PArallel Spatio-Temporal Attention with spatial auto-correlation gating (PASTA) that effectively captures the irregular spatio-temporal patterns of fine-grained maps. 
The novel components in our approach include spatial auto-correlation gating, multi-scale residual block, and temporal attention gating module. 
The spatial auto-correlation gating employs the concept of spatial statistics to identify irregular spatial regions. 
The multi-scale residual block is responsible for handling multiple range spatial dependencies in the fine-grained map, and the temporal attention gating filters out irrelevant temporal information for the prediction. 
The experimental results demonstrate that our model outperforms other competing baselines, especially under challenging conditions that contain irregular spatial regions. 
We also provide a qualitative analysis to derive the critical time information where our model assigns high attention scores in prediction.

\keywords{Fine-grained city-wide prediction \and Spatial auto-correlation \and Temporal attention module.}
\end{abstract}

\section{Introduction}
\;\;\;\;\; Location-based services with fine-grained maps are being widely introduced as a result of the continuing development of positioning technology \cite{lin2020preserving}.
These services provide city-wide crowd flow prediction to aid urban management, for use by the general public and policy makers \cite{zhu2020high,kapoor2020examining}. 
However, this prediction task turns out to be challenging, because the spatio-temporal dependencies are too complicated in the fine-grained map~\cite{tobler1970computer,liang2019urbanfm}.

\begin{figure}
    \includegraphics[width=0.7\linewidth]{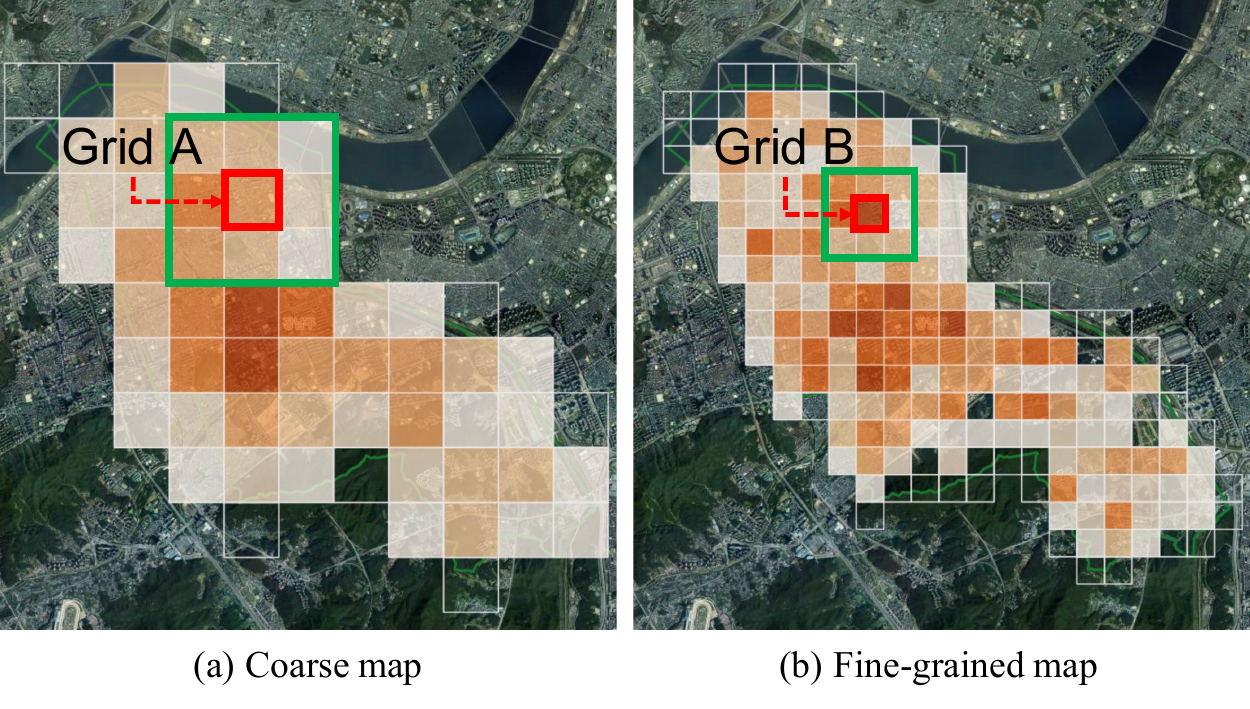}
    \caption{Illustration of the spatial distribution of crowd flows in coarse and fine-grained spatial maps. The larger the volume of crowd flows, the darker the color. The variance in the crowd flow in the fine-grained map is larger than in the coarse map. In the fine-grained map, grid\;$B$ includes a shopping complex where numerous people gather, and is a region of practical importance.}
    \label{fig:resolution}
    \end{figure}

Spatio-temporal prediction has been actively studied in many previous studies ~\cite{zhang2016deep, yao2019revisiting, lin2020preserving}. 
They partitioned a city into an $N \times M$ grid map based on latitude and longitude where the grid represents specific region. 
With this grid map, they predict the number of individuals (e.g., people or taxis) moving in each grid in the future. 
For example, ST-ResNet~\cite{zhang2016deep} proposed a method based on the residual convolution to predict crowd flows. 
In addition, STDN \cite{yao2019revisiting} designed a model of integrating convolution layers and long-short term memory (LSTM) to reflect both spatial and temporal dependencies jointly. 
The long-term prediction of spatio-temporal data has also been accomplished using a spatial-attention module \cite{lin2020preserving}.
However, these previous approaches are still inaccurate and inefficient in practice for predicting a fine-grained city-wide map because of the following three major factors:

\textbf{(1) Irregular spatial patterns}: According to the first law of geography, similar spatial attributes such as crowd flows have a tendency to be located near each other \cite{tobler1970computer}.
  However, as the spatial pattern becomes more dynamic and irregularly distributed as the resolution increases, cases that widely deviate from the first law of geography are often observed.
  Figure \ref{fig:resolution} describes the spatial distribution of the number of moving individuals of Seoul in South Korea.
  In the coarse map, a specific grid\;$(A)$ has a value similar to its neighbors. 
  However, grid\;$(B)$ in the fine-grained map is significantly different from its neighbors, creating spatially irregular patterns. 
  These irregular patterns are caused by a large value grid surrounded by small value grids (i.e., high-low), or vise-versa (i.e., low-high). 
  These high-low or low-high grids are difficult to predict and likely to be regions of practical importance, such as a key commercial hub.  
    
  Therefore, we apply a spatial normalization method that can reflect the spatially irregular patterns by leveraging the concept of local \textit{Moran's I} statistics representing  spatial auto-correlation \cite{anselin1995local}. 
  Spatial auto-correlation indicates the degree to which similar values are located spatially nearby.  
  The high-low and low-high grids have negative statistics with large absolute values in the local \textit{Moran's I} statistics. Therefore, \textit{Moran's I} statistics can be used to identify irregular spatial grids.
  
  \;\; \textbf{(2) Multi-range spatial dependency}: Another difficulty comes from multi-range spatial dependency in the fine-grained map.
  In other words, a specific grid has strong correlations from adjacent to distant grids in terms of connection distance.
  In addition, spatial distribution is naturally resolution-sensitive, and the interrelation of attributes varies according to the level of resolution \cite{zhang2014scale}. 
  For example, a fine-grained map may contain several geographical hierarchies such as districts and cities, and has multiple interrelations between regions.
  For this reason, a residual convolution-based architecture with multi-scale filters is adopted to effectively enlarge the receptive field of filters to incorporate the multi-range context in the fine-grained map.

  \;\; \textbf{(3) Irrelevant temporal information introducing noise}: In a historical observation, a considerable amount of irrelevant temporal information makes noise in the prediction.
  With a fine-grained map, this phenomenon becomes more pronounced, as the spatial pattern over time changes dynamically.
  Therefore, we introduce a temporal attention gating module \cite{woo2018cbam}, which weights core temporal information for the prediction.
  
  In short, to address these three issues, we propose a PArallel Spatio-Temporal Attention with spatial auto-correlation gating (PASTA) which consists of spatial auto-correlation gating (SAG) to reflect spatial auto-correlation, a multi-scale residual block (MSR) to handle multi-range spatial dependency, and a temporal attention gating module (TAG) to capture significant temporal features for predicting city-wide crowd flows. 
  Our model outperforms other competing baselines, especially under challenging conditions that contain irregular spatial regions.
  
\section{Related Work}  

\;\;\;\;\; Convolution-based models were proposed in many previous studies to predict future population crowds ~\cite{zhang2016deep,yao2019revisiting}.
Those convolution-based models learn spatial patterns which indicate high correlations between nearby regions.
An attention module is applied to filter out irrelevant information when predicting spatio-temporal data~\cite{yao2019revisiting,lin2020preserving}.

STDN \cite{yao2019revisiting} employs an attention module to capture long-term periodic information and temporal shifting.
DSAN \cite{lin2020preserving} implements long-term flow prediction with an attention module to minimize the impact of irrelevant spatial noise.

Previous studies also proposed a graph-based model to predict future crowd flows \cite{song2020spatial, guo2019attention}. 
For example, the spatio-temporal graph structure is employed, which connects all grids at the previous and next timestamps, to model spatial and temporal adjacency simultaneously \cite{song2020spatial}.
In addition, spatio-temporal attention, to capture the traffic network's dynamic spatial and temporal correlations, is proposed using an attention based spatial-temporal graph convolutional network \cite{guo2019attention}. 
However, we encounter several issues when predicting a fine-grained map, such as irregular spatial patterns.
For this reason, this study employs a module to reflect irregular spatial patterns using spatial statistics.
In addition, graph-based models are inefficient for fine-grained maps because of the computational cost required to process the large number of nodes in the spatio-temporal graph. 
Therefore, this paper uses residual convolution with temporal attention to efficiently train the model.

On the other hand, the spatio-temporal prediction task is similar to the semantic segmentation task which predicts a class label to every pixel in the image, in that input and output of the same size are used, and multi-range spatial dependency should be considered.
Therefore, to design a model suitable for our problem, we explored various architectures used in semantic segmentation task.
Among them, the DeepLab series \cite{chen2017deeplab, chen2017rethinking} encourages the active utilization of dilated convolution to solve the semantic segmentation task.
In particular, atrous spatial pyramid pooling is proposed to apply multi-scale context in DeepLab V2 \cite{chen2017deeplab} and DeepLab V3 \cite{chen2017rethinking}, which adopts parallel dilated convolutional filters to handle multi-scale objects.
 
In summary, most semantic segmentation tasks use the multi-filters strategy to cope with various scales. 
This study also devises a structure that uses convolution filters of different sizes in parallel to handle the multi-range spatial dependency.

\section{Proposed Method}
\subsection{Problem Definition}

\textbf{Problem setting.} This study partitions a city into $N \times M$ equal-sized grids based on latitude and longitude, where a grid represents a specific region. 
The grid is a rectangle corresponding to a small region in a specific city.
The actual grid resolution varies from 100m $\times$ 100m to 1km $\times$ 1km. 
We handle a fine-grained map whose grid map size is under 500m. Each grid contains flow volume with timestamps.

We use $x_{t}^{i,j} \in \mathbb{R}$ to denote the value of the $(i,j)$ grid ($i\in\{1,\cdots ,N\}$, $j\in\{1,\cdots ,M\}$) at timestamp $t$, and $X_{t} \in \mathbb{R}^{N \times M}$ denotes the values of all grid maps at timestamp $t$. 
We set $\mathcal{X}=(X_{t-T+1},X_{t-T+2},...,X_{t}) \in \mathbb{R}^{N \times M \times T}$ as the value of all grids over $T$ timestamps. 

$T$ timestamps consist of three fragments, denoting recent ($closeness$), daily-periodic ($periodic$), and weekly-periodic ($trend$) segments. 
Therefore, intervals of $T_{closeness}$, $T_{periodic}$, and $T_{trend}$ are an hour, a day, and a week respectively.
From this, the timestamps of $closeness$ refers to the timestamps immediately before the target timestamp to be predicted.
For example, if we predict the target timestamp $t$, then the timestamps of the $closeness$ in the input sequences can be $\{t-1, t-2, t-3\}$. 
The timestamps of $periodic$ means the identical daily past timestamps of the target timestamp and $trend$ refers to the same weekly past timestamps of the target timestamp.
The examples of the timestamps of $periodic$ and $closeness$ are $\{t-1\times24, t-2\times24, t-3\times24\}$ and $\{t-7\times24, t-14\times24\}$, respectively, given the target timestamp $t$.
The number of timestamps in each fragment is represented as $T_{k}$, where $k \in (closeness, period, trend)$ indicates each fragment. 
The total number of timestamps $T$ is equal to $T_{closeness}+T_{periodic}+T_{trend}$.

This study set $T_{closeness}$, $T_{periodic}$, and $T_{trend}$ as five, six, and four respectively. 
The $T_{k}$-channel crowd flows map of each time fragment is then concatenated with the channel axis.
In addition, $Y_{t+1}=X_{t+1} \in \mathbb{R}^{N \times M}$ represents the number of individuals in all grids to be predicted at timestamp $t+1$.

\textbf{Problem.} Given the historical sequence of all the grids over past $T$ slices $\mathcal{X}=(X_{t-T+1},X_{t-T+2},...,X_{t}) \in \mathbb{R}^{N \times M \times T}$, we aim to predict the future flows $Y_{t+1} \in \mathbb{R}^{N \times M}$.

    \begin{figure*}
    \includegraphics[width=1.0\linewidth]{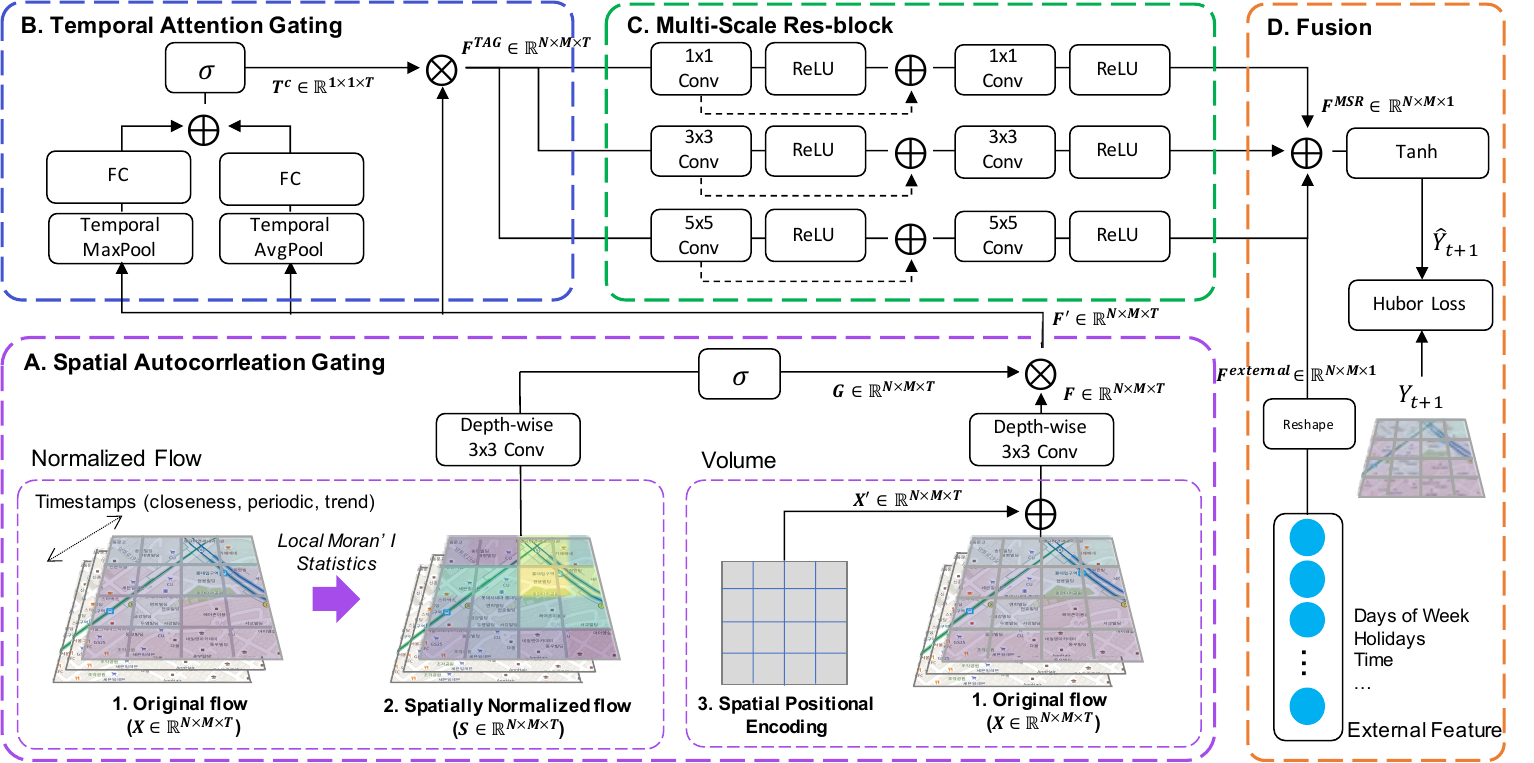}
    \caption{Architecture of PASTA.  There are three major components: spatial auto-correlation gating, temporal attention gating, and multi-scale residual block module. $FC$ indicates fully-connected layers and $Conv$ is a convolution filter. $\sigma$ is a sigmoid function. $\oplus$ denotes the element-wise summation, and $\otimes$ element-wise multiplication.}
    \label{fig:PASTA_total}
    \end{figure*}

\subsection{Parallel Spatio-Temporal Attention with spatial auto-correlation gating}
\subsubsection{Overview.}
Figure \ref{fig:PASTA_total} describes the architecture of PASTA, which is composed of three components including spatial auto-correlation gating (SAG), temporal attention gating (TAG), and multi-scale residual block (MSR) module.
First of all, we modify flow volumes throughout a city with $T$ timestamps into an $N \times M$ image-like matrix $\mathcal{X} \in \mathbb{R}^{N \times M \times T}$. 
We element-wise sum them with their corresponding spatial positional encodings (SPE) before entering the TAG module. 
The SPE is a particular bias to represent the location of grids \cite{lin2020preserving}.
The SAG module for local spatial normalization is applied to $\mathcal{X}$ to produce the indicator of the spatial auto-correlation $\mathcal{S} \in \mathbb{R}^{N \times M \times T}$.
Consequently, the normalized outputs $\mathcal{S}$ are fed into the depth-wise convolution layer and a sigmoid activation to produce gating values. 
The output of SAG is fed into the TAG module and then the MSR module. 
Such structure captures significant channel-wise features for temporal information and the multi-range dependency between nearby and distant regions.
In addition, the external features, such as weather, are fed into two-layer fully-connected neural networks  to extract latent features. 
These features are further element-wise summed with the outputs of the MSR. 
Lastly, a tanh activation function is adopted to map the aggregation into $[-1, 1]$. 
We adopts Huber loss \cite{huber1992robust} as the loss function, which is less sensitive to outliers than the mean squared loss.
\subsubsection{SPE: Spatial Positional Encoding.} 
Each grid has its own address. 
This is a characteristic that differentiates the location data from the image. 
To consider relative positions in the grid map, we equip the spatial positional encoding (SPE) as a bias to represent the position of a location \cite{lin2020preserving}.
Given the coordinate matrix, we calculate the SPE as follows

\begin{dmath}
SPE_{i,j}^{l}=\left \{\begin{array}{ll} sin(i/10000^{2l/d}) \;\;\; ,if \; l=2n,
     &  \\ cos(j/10000^{2l/d}) \;\;\; ,if \; l=2n+1,
\end{array} \right.
\end{dmath}
where $SPE_{i,j}^l \in \mathbb{R}$ is the $l$-th dimension element of encoding vector of position $(i,j)$ in the grid map. 
We element-wise sum the SPE encoding with the sequence of grid map $\mathcal{X}=(X_{t-T+1},X_{t-T+2},...,X_{t}) \in \mathbb{R}^{N \times M \times T}$ to produce $\mathcal{X'}=(X'_{t-T+1},X'_{t-T+2},...,\\X'_{t}) \in \mathbb{R}^{N \times M \times T}$.

\subsubsection{SAG: Spatial Auto-correlation Gating.}
In general, spatial information has the property of spatial auto-correlation, where spatial attributes such as crowd flows in nearby regions tend to be similar on a map. 
The spatial auto-correlation is denoted as the first law of geography~\cite{tobler1970computer}, which makes the proposition that \textit{"Everything is related with everything else, but near things are more related than distance things".}

However, as the map becomes finer, the opposite cases often appear, resulting in a large volume region surrounded by a small volume region (i.e., high-low), or vise-versa (i.e., low-high). 
These high-low or low-high region produce irregular spatial patterns contrary to the first law of geography.
Various indices have been proposed to measure the degree of spatial auto-correlation quantitatively.
In particular, \textit{Local Moran's I} statistic~\cite{anselin1995local} is a popular statistics to quantify local spatial auto-correlation. 

Suppose that $x^{i,j}_t$ is the value of the $(i,j)$ grid among overall grids in timestamp $t$, and $i$ and $j$ are the index of horizontal and vertical axis in the whole grid map respectively. 
In addition, $N$ and $M$ are the total number of grids in the horizontal and vertical axis of the map, respectively. 
Then the \textit{Local Moran's I} statistics $s_t^{i,j}$ of $(i,j)$ grid in timestamp $t$ is calculated as follows
\begin{dmath}
s_t^{i,j} = \frac{(x_t^{i,j}-\bar{x}_t)}{P_t}\sum_{z \in W_{ij}}{ \frac{(x_t^{z}-\bar{x}_t)}{P_{t}}}, 
\end{dmath}
where $\bar{x}_t=\frac{\sum_{i=1}^{N}\sum_{j=1}^{M}{x_t^{i,j}}}{NM}$ and  $P_t=\sqrt{\frac{\sum_{i=1}^{N}\sum_{j=1}^{M}{(x_t^{i,j}-\bar{x}_t)}}{NM-1}}$.
$W_{ij}$ is the set of predefined neighbor grids, which is all grids that share an edge or a corner with the $(i,j)$ grid.
Therefore, $x_t^{z}$ is the \textit{Local Moran's I} statistics of neighbor grids with the $(i,j)$ grid.

The statistics represent the relative value of a particular grid to its neighbors. 
If both a grid $(i,j)$ and its neighbors are greater than the overall average $\bar{x}_t$ together, then a positive $s_t^{i,j}$ is produced in the $(i,j)$ grid. 
Conversely, if the value of a grid $(i,j)$ is greater than $\bar{x}_t$ and its neighbors are less than $\bar{x}_t$, then the region $(i,j)$ may have a negative $s_t^{i,j}$ value. 
This region is denoted as high-low, and in the opposite case, low-high.
That is, a negative $s_t^{i,j}$ indicates that grid $(i,j)$ has a relatively large or small value compared to its neighbors.
These grids with negative $s_t^{i,j}$ are challenging to predict in that their spatial patterns are highly irregular.

The local \textit{Moran's I} statistics are calculated for all grids $\mathcal{X}$.
In Figure \ref{fig:PASTA_total}A, the illustrative calculation process of local \textit{Moran's I} statistics is described.
The output of local \textit{Moran's I} statistics $\mathcal{S} \in \mathbb{R}^{N \times M \times T}$ has the same shape as the original flow volumes $\mathcal{X}$. 
Then, the normalized output $\mathcal{S}$ is passed by depth-wise convolutional layer and sigmoid activation function to produce $\mathcal{G}=(G_{t-T+1},G_{t-T+2},...,G_{t}) \in \mathbb{R}^{N \times M \times T}$. 

Meanwhile, the original flow volume $\mathcal{X}$ is also applied to the depth-wise convolutional layer to extract the unique spatial pattern of each timestamp without sharing temporal information. 
It is represented as $\mathcal{F}=(F_{t-T+1},F_{t-T+2},...,F_{t}) \in \mathbb{R}^{N \times M \times T}$. 
Then, $\mathcal{G}$ are element-wise multiplied by $\mathcal{F}$ to produce a gated output $\mathcal{F'} \in \mathbb{R}^{N \times M \times T}$. 
This indicates that the normalized value $\mathcal{G}$ explicitly controls the volume information associated with neighbor regions.

    \begin{figure*}
    \includegraphics[width=1.0\linewidth]{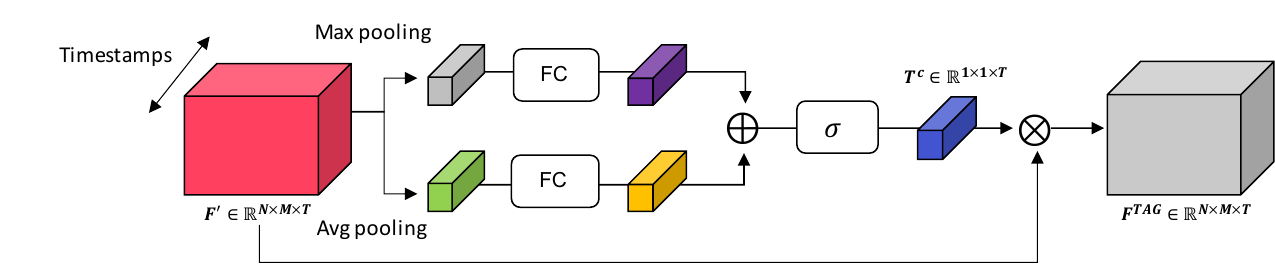}
    \caption{Diagram of the TAG module. 
    This module assigns high attention scores to meaningful timestamps of input sequences for prediction.
    Input $\mathcal{F'} \in \mathbb{R}^{N \times M \times T}$ (red tensor) is the output of the SAG module and the output $F^{TAG} \in \mathbb{R}^{N \times M \times T}$ (grey tensor), is fed into the MSR module. $FC$ indicates a fully-connected layer and $\sigma$ is a sigmoid function. 
    The element-wise summation and element-wise multiplication are denoted as $\oplus$ and $\otimes$, respectively.}
    \label{fig:TAG module}
    \end{figure*}

\subsubsection{TAG: Temporal Attention Gating.}
Some past historical information may be irrelevant for future prediction. 
For this reason, inspired by the channel attention module \cite{woo2018cbam}, we proposes the TAG module, which filters out temporal features that are irrelevant for prediction.
The TAG module is described in Figure \ref{fig:PASTA_total}B and Figure \ref{fig:TAG module}. 
It produces a temporal attention map to reflect the inter-time relationship. 
The channel indicates the timestamp.
This temporal attention focuses on the influential timestamps for prediction. 

Specifically, we implement the max-pooling and average-pooling operation to $\mathcal{F'}$, the outputs of SAG, and the both outputs $F_{max}^{c}$ and $F_{avg}^{c}$  are then forwarded to the two fully connected layers.
After the fully connected layer is applied to each feature, we merge them using element-wise summation with a sigmoid activation function. 
The temporal attention map $T^{c} \in \mathbb{R}^{1 \times 1 \times T}$ is described as

\begin{dmath}
T^{c} = \sigma(\text{FC}(\text{AvgPool}(\mathcal{F'}))+\text{FC}(\text{MaxPool}(\mathcal{F'})))
=\sigma (\textbf{W}_1\textbf{W}_0F_{avg}^{c} + \textbf{W}_3\textbf{W}_2F_{max}^{c}),
\end{dmath}
where $\sigma$ is the sigmoid function.
In addition, $\textbf{W}_0$,  $\textbf{W}_1$, $\textbf{W}_2$, and $\textbf{W}_3$ are the trainable weights in the fully connected layers.
Then, the final output of the TAG module,  $F^{TAG} \in \mathbb{R}^{N \times M \times T}$ is derived as follows

\begin{dmath}
F^{TAG}=T^{c} \otimes \mathcal{F'},
\end{dmath}
where $\otimes$ denotes element-wise multiplication.
Through this process, each channel is multiplied by weight between 0 and 1, so that the model can only take critical temporal information. 
A high attention weight will be given to timestamps essential for prediction.
\subsubsection{MSR: Multi-Scale Residual Block.}
It is crucial to consider multi-range spatial dependence when dealing with a fine-grained map. 
Therefore, we designed an MSR module composed of shallow layers using parallel convolution filters of different sizes.

The multiple filters are adopted with a parallel scheme as shown in Figure \ref{fig:PASTA_total}C.
The output feature map $F^{TAG} \in \mathbb{R}^{N \times M \times T}$ from the TAG module is fed into parallel convolutional layers with $1 \times 1$, $3 \times 3$, and $5 \times 5$ filters, respectively, followed by skip-connections. 
Then, the outputs are element-wise summed in time axis.
The input and output in the same shape (i.e., $N \times M$ ) are generated by using padding in a convolution layer.
The output feature map $F^{MSR} \in \mathbb{R}^{N \times M \times 1}$ is derived as follows

\begin{dmath}
F^{MSR}=\sigma(f^{1 \times 1}(\sigma(f^{1 \times 1}(F^{TAG}))+f^{1 \times 1}(F^{TAG}))+\sigma(f^{3 \times 3}(\sigma(f^{3 \times 3}(F^{TAG}))+f^{3 \times 3}(F^{TAG}))+\sigma(f^{5 \times 5}(\sigma(f^{5 \times 5}(F^{TAG}))+f^{5 \times 5}(F^{TAG})),
\end{dmath}
where $\sigma$ is the ReLU function and $f^{l \times l}$ indicates a convolution operation with the filter size $l \times l$.

\subsubsection{External Features.}
Numerous complex external factors significantly affect crowd flows.
We adopt time-of-day (24/48 dimensional variables for every 1 hour/30 minutes), day-of-week (7), and holiday (1) as the external features. 
The embeddings of external features from two fully-connected layers are concatenated with the outputs of the MSR module. 
In two full-connected layers of external features, the first layer is a fully-connected embedding layer followed by ReLU activation, and the second layer map low to high dimensions with the same shape as $F^{MSR}$.
This output is denoted as $F^{external}$. 
Then, $F^{MSR}$ and $F^{external}$ are summed in a channel-wise manner. 
Finally, the predicted value at the $t+1$ timestamp, denoted by $\hat Y_{t+1}$, is defined as

\begin{dmath}
\hat Y_{t+1}=\sigma(F^{MSR}+F^{external}),
\end{dmath}
where $\sigma$ is a tanh activation function.

\section{Experiment}

\subsection{Experiment Settings}
\textbf{Dataset.} We evaluated our model on two public real-world datasets from New York City (NYC). 
As shown in Table \ref{tab:data_public_1}, we used three large datasets, NYC-Taxi and NYC-Bike.
NYC-Taxi contains the taxi demands in $16$ $\times$ $12$ grids in New York City every 30 minutes from January 1, 2016 to February 29, 2016.  
The training set is set to data from January 1, 2016 to February 15, 2016, and the test set is set to the remaining data.
NYC-Bike contains the flows of bikes in $14$ $\times$ $8$ grids in New York City every 30 minutes from August 1, 2016 to September 29, 2016.  
We set the data from August 1, 2016 to September 15, 2016 as the training set and remaining data as the test set. 

In addition, we used fine-grained real-world data from the major cellular network operator in South Korea. 
This data is collected from the base stations. 
When users access the base stations for communication or data access, the logs including location records are generated in the base stations. 
The location data of about three million customers who agreed to collect and analyze their location information in Seoul were collected from April to June 2021. 
We set the data from April 1, 2021 to June 15, 2021 as the training set and the remaining data as the test set in Seoul-Crowd.
We normalized those datasets by min-max scale.

\begin{table}
  \label{tab:data_public_1}
     \caption{Summary of the dataset used in our experiments.}
  \vspace{-0.2cm}
  \begin{tabular}{{l c c c}}
    \hline\hline
    \textbf{Dataset} & \textbf{NYC-Taxi} & \textbf{NYC-Bike} & \textbf{Seoul-Crowd}\\
    \hline
    \ Data type & Taxi GPS & Bike rent & Mobile signal\\
    \ Location & New York & New York & Seoul\\
    \ Time span & 1/1-2/29, 2016 & 8/1-9/29, 2016 & 4/1-6/30, 2021\\
    \ Time interval & 30 mins  & 30 mins & 1 hour\\
    \ Grid map size & (16, 12)  & (14, 8) & (68, 92)\\
    \hline\hline
\end{tabular}
\end{table}

\noindent\textbf{Compared Algorithms.} To evaluate the accuracy of our predictive model, we compared the proposed model with several competitive methods:
\begin{itemize}
   \setlength{\itemsep}{0pt}
   \item ST-ResNet\cite{zhang2016deep}: It consists of simple architectures with residual block with CNN for spatio-temporal data.
   \item STDN\cite{yao2019revisiting}: This research is based on the local CNN and LSTM to capture the complex spatial dependencies and temporal dynamics.
  \item DSAN\cite{lin2020preserving}: They focus on the long-term prediction task with an attention module to deal with the dynamic correlation of spatio-temporal data. It is a state-of-the-art model in NYC-Taxi and NYC-Bike dataset.
\end{itemize}

\subsection{Results}
\subsubsection{Fine-grained map prediction.}
We experimented with fine-grained crowd flow data as shown in Table \ref{tab:resolution_result}.
We measured performance for resolutions of $12 \times 16$, $24 \times 32$, $48 \times 64$, and $68 \times 92$. 
Typically, the higher the resolution, the lower the model performance. 
This is because the finer the map, the more spatio-temporal irregularity occurs.
However, our model showed robust performance in the fine-grained map with $68 \times 92$ resolution and better performance than other models.
From the result, we discovered that reflecting spatial auto-correlation and multi-range spatial dependencies in the model can reduce error rate in fine-grained map prediction. 
In addition, we determined that filtering out irrelevant temporal information was critical to achieving reliable performance.

In order to better understand the effectiveness of our model, we carried out additional experiments as shown in Table \ref{tab:spatial_irregular_experiment_}.
In this experiment, we used our Seoul-Crowd dataset of $68 \times 92$. 
We observed that our model outperformed other models in high-low and low-high grids. 
These results demonstrate that our model makes more robust predictions for spatially irregular regions than other competitive baselines. 
This is because spatially irregular regions are discriminated in our model, using our SAG module, to prevent smoothing of the predicted values by surrounding regions.
\begin{table}
\caption{Fine-grained map predicion results (Seoul-Crowd)}
  \vspace{-0.7cm}
\label{tab:resolution_result}
\begin{tabular}{c|cc|cc|cc|cc}
\noalign{\smallskip}\noalign{\smallskip}\noalign{\smallskip}\noalign{\smallskip}\hline\hline
 Resolution & \multicolumn{2}{c|}{12 $\times$ 16} & \multicolumn{2}{c|}{24 $\times$ 32} & \multicolumn{2}{c|}{48 $\times$ 64} & \multicolumn{2}{c}{68 $\times$ 92} \\
\cline{2-9}
      & RMSE  & MAPE & RMSE & MAPE & RMSE & MAPE & RMSE & MAPE \\
\hline
 ST-ResNet\cite{zhang2016deep} & 115.94 & 22.47 & 114.43 & 22.23 & 118.11 & 21.73 & 132.20 & 23.92 \\
 STDN\cite{yao2019revisiting} & 101.11 & 17.85 & 97.43 & 17.22 & 114.53 & 17.78 & 119.30 & 21.67 \\
 DSAN\cite{lin2020preserving} & 98.12 & 16.11 & 93.43 & 15.98 & 99.10 & 16.22 & 113.44 & 20.22 \\
\hline
 \textbf{PASTA} & \textbf{96.57} & \textbf{15.49} & \textbf{91.79} & \textbf{14.55} & \textbf{96.21} & \textbf{14.58} & \textbf{108.55} & \textbf{18.92} \\
\hline
\hline
\end{tabular}
\end{table}

\subsubsection{Coarse map prediction.}
We evaluated the effectiveness of PASTA and other baselines using a coarse map dataset.
Table \ref{tab:crowds_result} shows the baseline performances. 
In some cases, our model significantly outperformed other baselines, achieving the lowest RMSE or MAPE even on the coarse map dataset. 
\begin{table}
\centering
\caption{Result of spatial irregular regions prediction with fine-grained map}
  \vspace{-0.5cm}
\label{tab:spatial_irregular_experiment_}
\medskip
\begin{tabular}{ccccc}
\hline\hline
Model & \multicolumn{2}{c}{High-Low} & \multicolumn{2}{c}{Low-High} \\ \cline{2-5} 
                      & RMSE     & MAPE     & RMSE     & MAPE     \\ \hline
ST-ResNet\cite{zhang2016deep}              & 139.77            & 33.65             & 60.76             & 26.88             \\
STDN\cite{yao2019revisiting}                   & 135.98            & 31.10             & 57.73             & 25.08             \\
DSAN\cite{lin2020preserving}                   & 135.33            & 30.41             & 58.81             & 25.31             \\ \hline
\textbf{PASTA}                  & \textbf{132.12}            & \textbf{28.78}             & \textbf{56.12}             & \textbf{24.78}             \\ \hline\hline
\end{tabular}
\end{table}

\begin{table}
\caption{Coarse map prediction results}
  \vspace{-0.2cm}
 \label{tab:crowds_result}
\begin{tabular}{cc|cc|cc|cc}
\hline\hline
Model  & Dataset  & \multicolumn{2}{c|}{NYC-Taxi} & \multicolumn{2}{c|}{NYC-Bike} & \multicolumn{2}{c}{Crowd-Flow} \\ \cline{2-8} 
                & Metric   & RMSE      & MAPE     & RMSE     & MAPE      & RMSE      & MAPE      \\ \hline
\multicolumn{2}{c|}{ST-ResNet\cite{zhang2016deep}}      & 23.82              & 18.87             & 9.43              & 22.03              & 115.94             & 22.47              \\
\multicolumn{2}{c|}{STDN\cite{yao2019revisiting}}           & 22.98              & 17.88             & 9.41              & 19.94              & 101.11             & 17.83              \\
\multicolumn{2}{c|}{DSAN\cite{lin2020preserving}}           & 20.73              & \textbf{16.09}    & \textbf{8.03}     & 18.33              & 98.12              & 16.17              \\
\hline
\multicolumn{2}{c|}{\textbf{PASTA}} & \textbf{19.89}     & 16.12             & 8.26              & \textbf{17.76}     & \textbf{96.57}     & \textbf{15.49}     \\ \hline\hline
\end{tabular}
\end{table}

\subsection{Ablation study}
\;\;\;\;\; 
We investigated the effectiveness of each module of our model.
Table \ref{tab:ablation_study} shows the experimental results, depending on whether each module was applied or not.
From Rows (C) and (F), we observe that the SAG module improved the performance of our model. 
This result indicates that the SAG module helps improve the robustness in the fine-grained map prediction by correcting the spatially irregular pattern.
Rows (D) and (F) show the effect of the MSR module. 
We found that the model with the MSR module was better than the model without it. 
This is because the MSR module allows the model to cope with multi-scale spatial dependency.
The effect of the TAG module is illustrated in Rows (E) and (F). 
This result demonstrates that filtering out irrelevant temporal feature is indispensable to derive more accurate prediction.

\begin{table}
\centering
\caption{Component analysis }
  \vspace{-0.3cm}
\label{tab:ablation_study}
\begin{tabular}{cccc|cc}
\hline\hline
Model & SAG & TAG & MSR & RMSE & MAPE \\ \hline
(A) &   & $\checkmark$ &    &   116.12  &   24.11   \\
(B) &   &    & $\checkmark$ &   117.30  &   24.98   \\
(C) &   & $\checkmark$ & $\checkmark$ &  114.43  &   21.28   \\
(D) & $\checkmark$ & $\checkmark$ &   &  113.12  & 21.09     \\
(E) & $\checkmark$ &   & $\checkmark$ & 113.56  & 21.14     \\
\hline
(F) & $\checkmark$ & $\checkmark$ & $\checkmark$ &  \textbf{108.55}    &  \textbf{18.92}    \\ \hline\hline
\end{tabular}
\end{table}

\subsection{Visualization of Temporal Attention Gating}
\begin{figure}
    \includegraphics[width=1\linewidth]{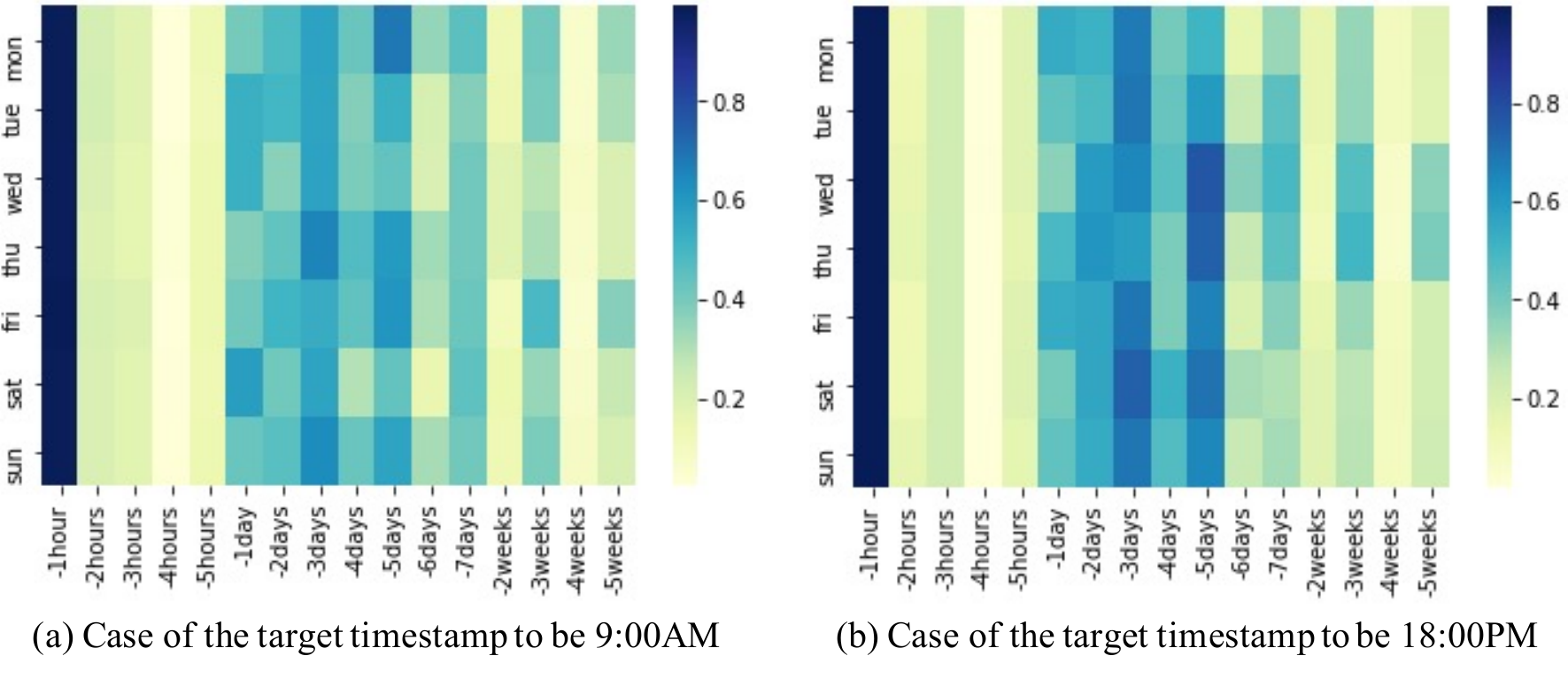}
    \caption{An illustrative example of temporal attention maps. The higher the attention weights, the darker the color. The y-axis indicates the day of the week and the x-axis is the timestamp. Each grid denotes the average attention weights of the input. (a) is an attention map where the target timestamp to be predicted is 9:00 AM. (b) is the attention map where the target timestamp to be predicted is 18:00 PM.}
    \label{fig:time_attentionamp}
    \end{figure}

\;\;\;\;\; We visualized the temporal attention map $T^{c} \in \mathbb{R}^{1 \times 1 \times T}$ of the TAG module to verify the weight of temporal information.
Each input sequence $\mathcal{X} \in \mathbb{R}^{N \times M \times T}$ produces $T^{c}$ respectively.
We averaged the temporal attention map $T^{c}$ from our test dataset of Seoul-Crowd for the $T$ timestamps.
As shown in Figure \ref{fig:time_attentionamp}, the model assigns large weights to the feature map of an hour ago. 
This shows that the timestamp of an hour ago is critical to the prediction.
However, if the timestamp is close to the target timestamp but not the same period, low weights are given to that timestamp. 
The attention weights of 2,3,4, and 5 hours ago are below 0.3. 
In addition, the model sometimes gives low weights for feature maps in the far distant past, where the attention weights of 2 and 4 weeks ago are below 0.3. 
This visualization demonstrates that not all time information is required for prediction.

\section{Conclusion}

\;\;\;\;\; In this work, we proposed PASTA for predicting future city-wide crowd flows.
Our model consists of a SAG module to reflect spatial auto-correlation, an MSR module to handle multi-range spatial dependency and a TAG module to filter out irrelevant temporal features for prediction. 
This study discovered that reflecting spatial relativity and multi-range spatial dependencies in the model can reduce error rates in fine-grained map prediction. 
In addition, we determined that filtering out irrelevant temporal information was critical to achieving reliable performance.
We extensively evaluated our model on future flow prediction tasks using real-world datasets of fine-grained maps. 
Our model outperformed other competing baselines in both fine-grained and coarse maps cases.
In addition, the results also showed that our model performed better predicting regions with spatially irregular patterns.

\subsubsection{Acknowledgment}
This work was supported by Institute of Information $\&$ communications Technology Planning $\&$ Evaluation (IITP) grant funded by the Korea government(MSIT) (No. 2020-0-00368, A Neural-Symbolic Model for Knowledge Acquisition and Inference Techniques, No.2019-0-00075, Artificial Intelligence Graduate School Program(KAIST)).

\noindent The authors would like to thank T3K center of SK Telecom for providing GPU cluster support to conduct massive experiments.

\bibliographystyle{splncs04}
\bibliography{ref}

\end{document}